\documentclass{article}

\usepackage{arxiv}

\usepackage[utf8]{inputenc} % allow utf-8 input
\usepackage[T1]{fontenc}    % use 8-bit T1 fonts
\usepackage{hyperref}       % hyperlinks
\usepackage{url}            % simple URL typesetting
\usepackage{booktabs}       % professional-quality tables
\usepackage{amsfonts}       % blackboard math symbols
\usepackage{amsmath}        % math environments and commands
\usepackage{nicefrac}       % compact symbols for 1/2, etc.
\usepackage{microtype}      % microtypography
\usepackage{lipsum}         % Can be removed after putting your text content
\usepackage{graphicx}
\usepackage{xcolor}        % for colored text
\usepackage{colortbl}      % for colored table cells
\usepackage{cleveref}      % for \cref command
\usepackage{enumitem}      % for list customization
\usepackage{multirow}      % for multirow in tables
\usepackage{subcaption}    % for subfigures
\usepackage{natbib}
\usepackage{doi}

\title{FoldAct: Efficient and Stable Context Folding for Long-Horizon Search Agents}

\renewcommand{\shorttitle}{FoldAct: Efficient and Stable Context Folding for Long-Horizon Search Agents}

\author{
  Jiaqi Shao \\
  Hong Kong University of Science and Technology, Hong Kong \\
  Duke Kunshan University, China \\
  \And
  Yufeng Miao \\
  Microsoft AI\\
  \And
  Wei Zhang \\
  Hong Kong University of Science and Technology, Hong Kong \\
  \And
  Bing Luo \\
  Duke Kunshan University, China \\
}

\hypersetup{
pdftitle={FoldAct: Efficient and Stable Context Folding for Long-Horizon Search Agents},
pdfsubject={Machine Learning, Reinforcement Learning, Large Language Models},
pdfauthor={Author Names},
pdfkeywords={Machine Learning, Reinforcement Learning, Large Language Models, Context Folding, Long-Horizon Agents},
}

\begin{document}
\maketitle

% Main content (abstract is in main_body.tex)

\begin{abstract}
  Long-horizon reinforcement learning (RL) for large language models faces critical scalability challenges from unbounded context growth, leading to context folding methods that compress interaction history during task execution.
  However, existing approaches treat summary actions as standard actions, overlooking that summaries fundamentally modify the agent's future observation space, creating a policy-dependent, non-stationary observation distribution that violates core RL assumptions.
  This introduces three fundamental challenges: (1) gradient dilution where summary tokens receive insufficient training signal, (2) self-conditioning where policy updates change summary distributions, creating a vicious cycle of training collapse, and (3) computational cost from processing unique contexts at each turn.
  We introduce \textbf{FoldAct}\footnote{https://github.com/SHAO-Jiaqi757/FoldAct}, a framework that explicitly addresses these challenges through three key innovations: separated loss computation for independent gradient signals on summary and action tokens, full context consistency loss to reduce distribution shift, and selective segment training to reduce computational cost.
  Our method enables stable training of long-horizon search agents with context folding, addressing the non-stationary observation problem while improving training efficiency with 5.19$\times$ speedup.
\end{abstract}
\section{Introduction}

\begin{figure*}[t]
  \centering
  \includegraphics[width=0.8\textwidth]{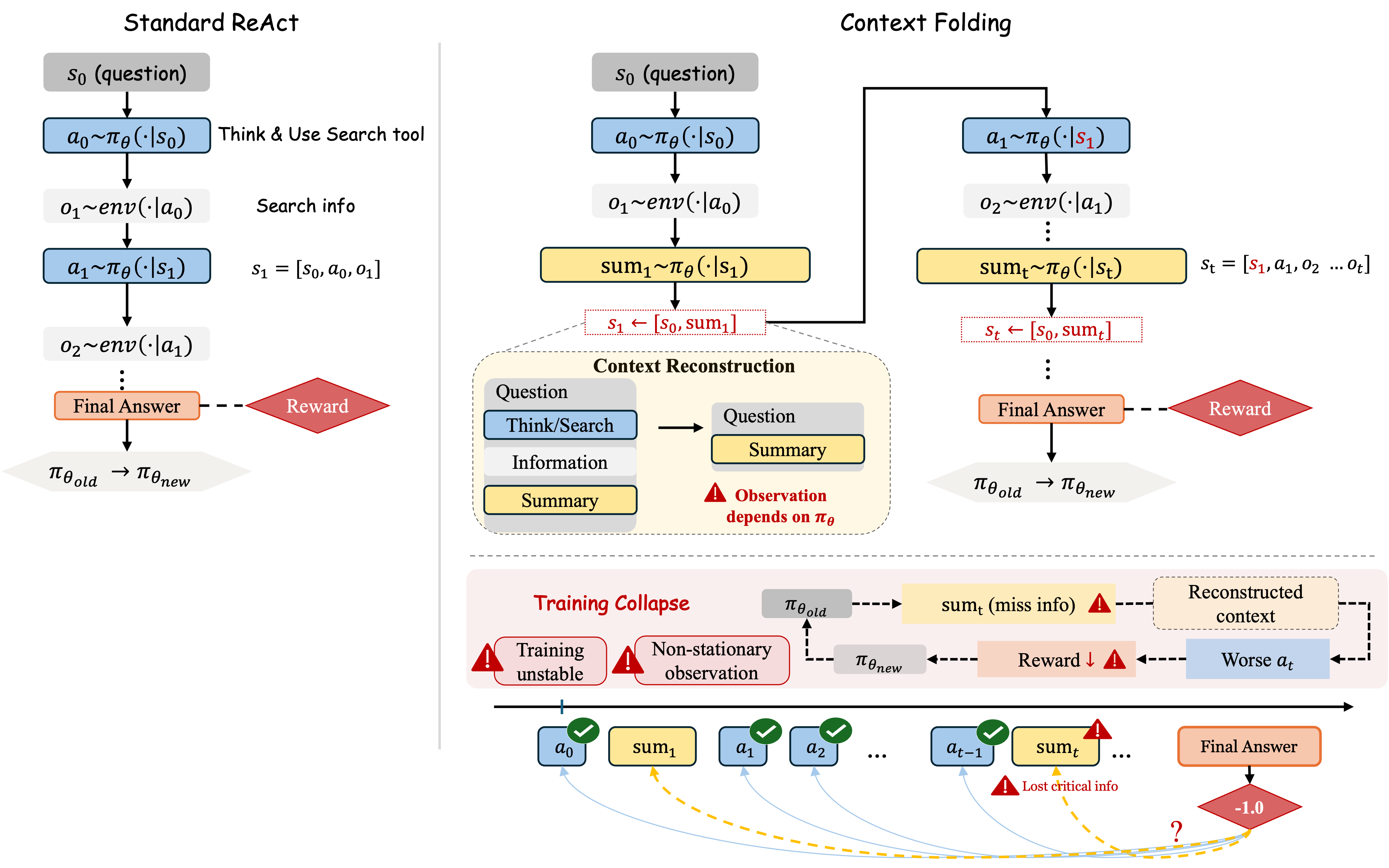}
  \caption{Comparison between standard RL (left) and RL with context folding (right). 
  \textbf{Left:} Standard RL assumes stationary observations independent of the policy. 
  \textbf{Right Top:} In context folding, summaries $\text{sum}_t \sim \pi_\theta(\cdot|s_t)$ are generated by the policy itself, and the visible state $s_t = [s_0, \text{sum}_t]$ is reconstructed from summaries, creating a self-conditioning loop where observations depend on $\pi_\theta$ (red warning). 
  \textbf{Right Bottom:} This leads to training collapse: poor summaries from $\pi_{\theta_{\text{old}}}$ miss critical information, causing reward degradation and further policy degradation in a vicious cycle of non-stationary observations and unstable training.
  }
  \label{fig:context_folding_overview}
\end{figure*}

Long-horizon reinforcement learning (RL) for large language models (LLMs) enables agents to solve complex multi-step tasks through sequential decision-making~\cite{chen2025reinforcement,xi2025agentgym}, but faces two fundamental scalability challenges: (1) \textbf{growing context} during inference as agents accumulate observations and actions over long trajectories~\cite{lu2025scalingllmmultiturnrl}, and (2) \textbf{computational bottlenecks} during training where rollouts require processing increasingly long contexts, dramatically slowing down the learning process~\cite{sun2025scaling,tan2025earl}.

Existing approaches to context folding fall into two categories: (1) \textbf{End-to-end methods} that incorporate folding into the RL framework (e.g., \citet{sun2025scaling} and \citet{ye2025agentfold} treat folding as function calls in the action space), and (2) \textbf{Separate models} that treat folding as external preprocessing~\cite{wan2025compass} and provide feedback signals~\cite{sun2025scaling,wu2025resum}.

However, both approaches fail to recognize that summary actions fundamentally 
differ from standard actions (e.g., tool calls or reasoning steps), 
overlooking that summary actions directly modify the agent's \textbf{future 
observation space}: they reconstruct what context the agent will observe in 
subsequent turns. 
This creates a fundamental problem where the observation distribution becomes policy-dependent.
Consequently, policy updates will also 
change summary action, creating a \textbf{non-stationary observation 
distribution}. 
The distinct role of summary tokens introduces three key challenges, detailed below:

\textbf{C1: Credit Assignment.} Standard RL computes a unified gradient over all tokens, treating summary and action tokens identically. This causes \textbf{gradient dilution}: if summary tokens constitute only 10\% of total tokens, they receive only 10\% of the gradient signal, even though folding decisions are critical for long-horizon performance. This prevents proper credit attribution between folding decisions and task execution actions.

\textbf{C2: Self-Conditioning.} Since summaries become part of future observations, policy updates change summary distributions, which in turn change observation distributions. This creates a vicious cycle: poor summaries from $\pi_{\theta_{\text{old}}}$ miss critical information, causing reward degradation, which further degrades the updated policy $\pi_{\theta_{\text{new}}}$, leading to training collapse (illustrated in \Cref{fig:context_folding_overview}, \textit{Right Bottom}).

\textbf{C3: Training Cost.} To ensure log-probability consistency between training and inference, we must train on the actual context $s_t$ used at each turn during rollout. Since each turn may observe a different compressed context (after summary generation, \Cref{fig:context_folding_overview}, \textit{Right Top}), training requires separate trajectory segments~\cite{wu2025resum}, making training computationally expensive.

\begin{figure*}[t]
  \centering
  \includegraphics[width=0.8\textwidth]{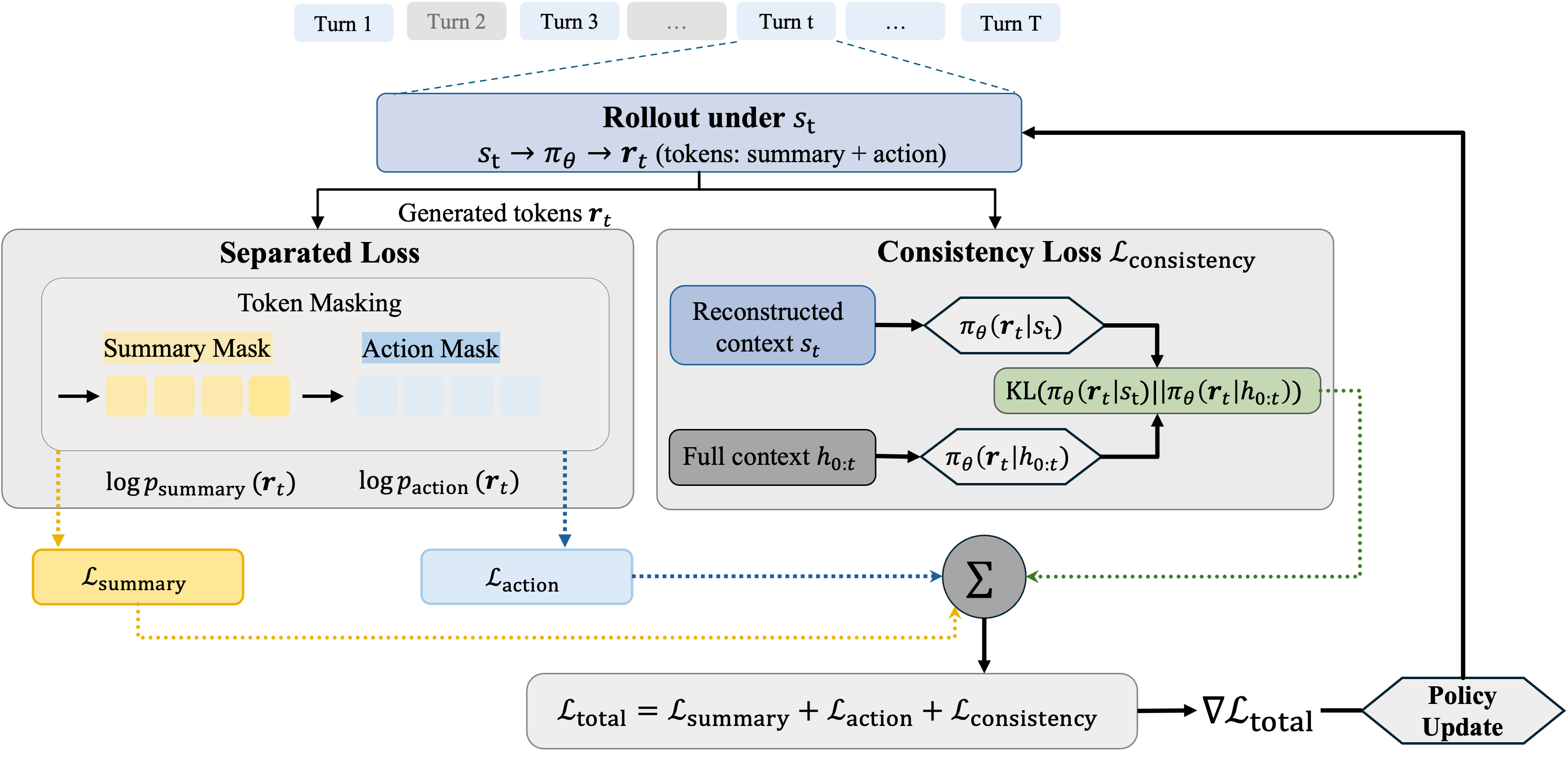}
  \caption{ Our solution framework addresses all three challenges through three key components: (1) Separated loss computation (yellow/blue paths) computes independent gradients for summary and action tokens ($\mathcal{L}_{\text{summary}}$ and $\mathcal{L}_{\text{action}}$), preventing gradient dilution and enabling correct credit assignment (C1). (2) Consistency loss (green path) minimizes KL divergence between compressed context $s_t$ (used during rollout) and full context $h_{0:t}$ (stored offline), reducing distribution shift from self-conditioning (C2). (3) Selective segment training trains only on sampled turns per trajectory (C3).
  }
  \label{fig:solution_architecture}
\end{figure*}

To tackle these challenges, we introduce \textbf{FoldAct} (see \Cref{fig:solution_architecture}), a framework that explicitly addresses the non-stationary, policy-dependent nature of summary actions in long-horizon RL. 
\textbf{FoldAct} consists of three key contributions for addressing C1-C3: 
\begin{enumerate}[noitemsep,topsep=0pt,parsep=0pt,partopsep=0pt,leftmargin=*]
    \item \textbf{Separated loss computation} (C1): We compute independent gradients for summary and action tokens, reducing gradient dilution and enabling proper credit attribution between folding and action decisions.
    \item \textbf{Full context consistency loss} (C2): We minimize the KL divergence between policy distributions conditioned on summarized and full contexts, thereby reducing distribution shifts and alleviating the self-conditioning problem.
    \item \textbf{Selective segment training} (C3): Instead of computing the loss at every turn, we sample a subset of trajectory turns for training, substantially lowering computational costs without sacrificing performance.
\end{enumerate}

\section{Related Work}

\subsection{Long-Horizon RL Agents with Context Constraints}

Long-horizon RL agents have been successfully applied to multi-turn tool-use tasks~\cite{xi2025agentgym} and information seeking~\cite{searchr1,asearcher,wu2025webdancer,li2025websailor}, where agents interact with external tools, functions, or environments over extended trajectories.
The ReAct paradigm~\cite{yao2022react}, adopted by contemporary web agents, employs an iterative reasoning-action-observation loop that accumulates all interactions in an append-only context.
However, this design leads to \textbf{context saturation} on long-horizon tasks, where critical signals become buried in noise, impairing reasoning capabilities.
Existing RL methods for training LLMs on multi-turn tasks are largely limited by fixed context lengths~\cite{lu2025scalingllmmultiturnrl}, which fundamentally bounds the difficulty of solvable tasks and prevents scaling to longer horizons. 
\subsection{Context Management in Long-Horizon RL} 
Long-horizon RL agents face unbounded context growth during task execution, necessitating context management strategies.
While external memory approaches~\cite{yan2025memory} learn memory operations for external storage and retrieval, our work focuses on \textbf{compressing working contexts} through LLM-based summarization, which directly reduces the active context size rather than augmenting it with external storage.
Existing compression methods progress from heuristic fixed rules (not optimized end-to-end), to rigid step-wise methods~\cite{zhou2025mem1,yu2025memagent} that compress full history at each step using fixed policies but are limited to simple retrieval tasks, to flexible methods like AgentFold~\cite{ye2025agentfold} that enable multi-scale selective compression through flexible look-back mechanisms with supervised fine-tuning, to end-to-end RL methods like SUPO~\cite{lu2025scalingllmmultiturnrl} and FoldAgent~\cite{sun2025scaling} that incorporate summarization into the RL framework and optimize jointly with task actions. Related approaches include ACON~\cite{kangACONOptimizingContext2025} which optimizes context compression, and IterResearch~\cite{chen2025iterresearchrethinkinglonghorizonagents} which addresses state reconstruction in long-horizon agents.
Critically, all methods treat summaries as standard actions, overlooking that summaries directly modify the agent's future observation space, violating the stationary observation assumption fundamental to policy gradient methods and creating non-stationary observation distributions.

\section{Method}
\label{sec:method}

\subsection{Problem Formulation}

We formalize context folding where summary generation fundamentally differs from standard actions. Unlike standard RL where observations are environment-provided and stationary, in our setting, summaries generated by the policy $\pi_\theta$ directly modify the agent's future observation space, creating a self-conditioning loop that violates fundamental RL assumptions.

\textbf{Problem Setup.} We model a long-horizon language agent operating in a ReAct-style multi-turn dialogue. The agent's trajectory follows:
\begin{equation}
s_0 \rightarrow a_0 \rightarrow o_1 \rightarrow \ldots \rightarrow \text{sum}_t \rightarrow a_t \rightarrow o_{t+1} \rightarrow \ldots
\label{eq:trajectory}
\end{equation}
where $s_0$ is the initial question, $a_t$ are actions (e.g., search, API calls), $o_t$ are environment observations, the agent generates summaries $\text{sum}_t$ and reconstructs the context as $s_t = [s_0, \text{sum}_t]$, as illustrated in~\Cref{fig:context_folding_overview} (\textit{Right Top}).

\textbf{Summary and Action Generation.} 
The agent decides whether to summary at turn $t$: $\text{sum}_t \sim \pi_\theta(\text{sum}_t | s_t)$. At every turn (including after summary generation), the agent generates actions: $a_t \sim \pi_\theta(a_t | s_t)$. Both summary and action generation use the same policy model $\pi_\theta$.

\textbf{The Self-Conditioning Problem.} As illustrated in \Cref{fig:context_folding_overview}, standard RL assumes \textbf{stationary observations} that independent of the policy (\Cref{fig:context_folding_overview}, \textit{Left}). However, when summaries are generated by $\pi_\theta$, we have:
\begin{equation}
% \text{sum}_t \sim \pi_\theta(\text{sum}_t | s_t), \quad
s_t \leftarrow [s_0, \text{sum}_t], \quad a_t \sim \pi_\theta(a_t | s_t)
\label{eq:selfconditioning}
\end{equation}
The policy updates change summary generation $\pi_\theta(\text{sum}_t | s_t)$, which change observation distributions $p(s_t | \pi_\theta)$, violating the stationary observation assumption. 
As shown in~\Cref{fig:context_folding_overview} (\textit{Right Bottom}), this leads to \textbf{training collapse}: poor summaries from $\pi_{\theta_{\text{old}}}$ miss critical information, causing reward degradation, which further degrades the updated policy $\pi_{\theta_{\text{new}}}$, creating a vicious cycle of non-stationary observations and unstable training.

To address the three fundamental challenges arising from this framework, we propose: (1) \textbf{Separated loss computation} computes independent gradients for summary tokens $\mathcal{L}_{\text{summary}}$ and action tokens $\mathcal{L}_{\text{action}}$, enabling proper credit assignment between folding and task execution decisions (Challenge 1). (2) \textbf{Full context consistency loss} minimizes KL divergence $D_{\text{KL}}(\pi_\theta(\cdot|s_t) || \pi_\theta(\cdot|h_{0:t}))$ between policy distributions under compressed context $s_t$ (used during rollout) and full context $h_{0:t}$ (stored offline), reducing distribution shift from self-conditioning (Challenge 2). (3) \textbf{Selective segment training} trains only on sampled turns per trajectory. 
The complete framework is illustrated in \Cref{fig:solution_architecture}.

\begin{figure*}[t]
  \centering
  \includegraphics[width=0.8\textwidth]{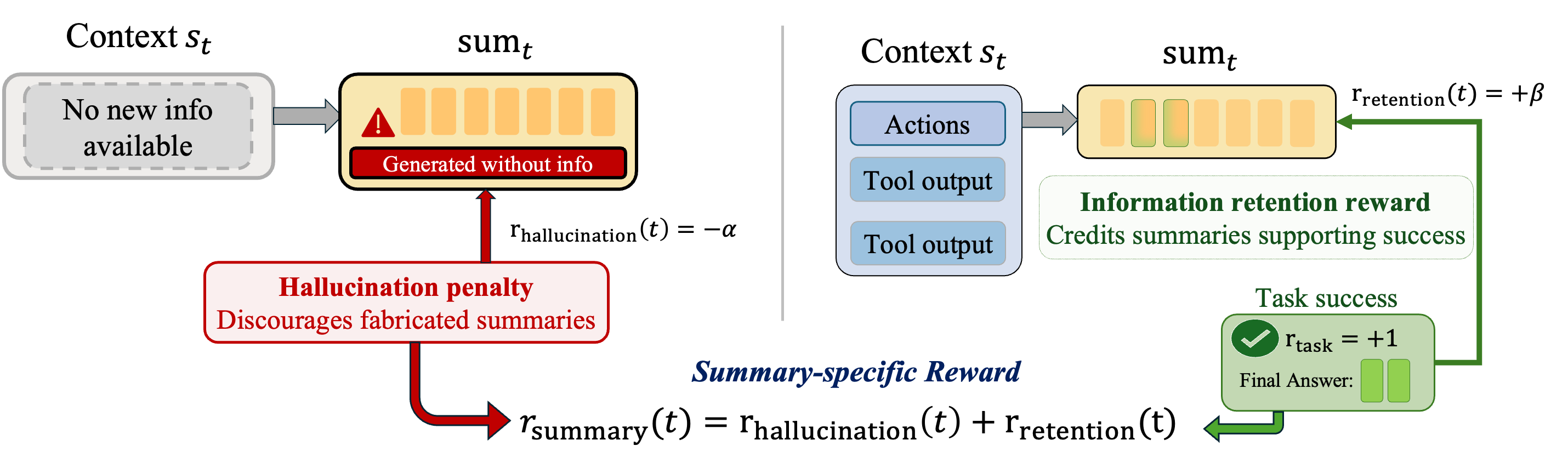}
  \caption{Summary Reward Design. The summary reward $r_{\text{summary}}(t)$ consists of two components: (1) \textbf{Hallucination Penalty} $r_{\text{hallucination}}(t)$ penalizes summaries generated when no information is available, preventing fabricated content. (2) \textbf{Information Retention Reward} $r_{\text{retention}}(t)$ rewards summaries that preserve critical information leading to task success. This design provides direct training signals for summary generation, enabling the agent to learn which information to retain in compressed contexts.}
  \label{fig:summary_reward_design}
  \end{figure*}
\subsection{Separated Loss Computation}

Standard policy gradient methods compute a unified gradient over all tokens, treating summary and action tokens identically. This creates a \textbf{gradient dilution problem}: if summary tokens constitute only 10\% of total tokens, they receive only 10\% of the gradient signal, even though folding decisions are critical for long-horizon performance. Moreover, when the agent fails due to missing information, standard policy gradients cannot properly attribute blame between (1) poor summary generation that discarded critical information, and (2) poor action selection given incomplete context.

\textbf{The Gradient Dilution Problem.} At the token level, the response sequence at turn $t$ contains both summary tokens $\mathbf{r}_t^{\text{summary}}$ and action tokens $\mathbf{r}_t^{\text{action}}$:
\begin{equation}
\mathbf{r}_t = [\mathbf{r}_t^{\text{summary}}, \mathbf{r}_t^{\text{action}}]
\label{eq:token_decomposition}
\end{equation}
Standard policy gradient methods compute a unified gradient over all tokens. From the sequence probability decomposition $\log \pi_\theta(\mathbf{r}_t | s_t) = \sum_{i=1}^{|\mathbf{r}_t|} \log \pi_\theta(r_{t,i} | \mathbf{r}_{t,<i}, s_t)$, the policy gradient is:
\begin{equation}
\nabla_\theta J(\theta) = \mathbb{E}\left[
\hat{A}_t \sum_{i=1}^{|\mathbf{r}_t|} \nabla_\theta \log \pi_\theta(r_{t,i} | \mathbf{r}_{t,<i}, s_t)
\right]
\label{eq:standard_token_loss}
\end{equation}
where $\hat{A}_t$ is the advantage estimate. When summary tokens are few (e.g., 10\% of total tokens), their gradient contribution is diluted: summary tokens receive approximately $\frac{|\mathbf{r}_t^{\text{summary}}|}{|\mathbf{r}_t|} \approx 0.1$ of the total gradient magnitude, even though folding decisions are critical for long-horizon performance.
This leads to summary generation receiving insufficient training signal for optimizing summary strategies.

% Decomposing the gradient by token category:
% \begin{equation}
% \nabla_\theta J(\theta) = \mathbb{E}\left[
% \hat{A}_t \left(
% \sum_{i \in \text{summary}} \nabla_\theta \log \pi_\theta(r_{t,i} | \mathbf{r}_{t,<i}, s_t)
% +
% \sum_{j \in \text{action}} \nabla_\theta \log \pi_\theta(r_{t,j} | \mathbf{r}_{t,<j}, s_t)
% \right)
% \right]
% \label{eq:gradient_decomposition}
% \end{equation}

\textbf{Separated Gradient Computation.} To address this, we compute independent gradients for summary and action tokens using token masks. In practice, these gradients are implemented via category-specific surrogate losses. We define a binary mask $\mathbf{m}_t$ that identifies summary tokens:
\begin{equation}
  \mathbf{m}_t^{\text{sum}}[i] =
  \begin{cases}
  1, & \text{if } r_{t,i} \text{ is a summary token}, \\
  0, & \text{otherwise}.
  \end{cases}
  \label{eq:summary_mask}
  \end{equation}
The mask for action tokens is defined as
  \(\mathbf{m}_t^{\text{act}} = 1 - \mathbf{m}_t^{\text{sum}}\).
More generally, we use $c \in \{\text{sum}, \text{act}\}$ to denote the token category,
with the corresponding mask $\mathbf{m}_t^{(c)}$.
% Given policy $\pi_\theta$, the log-probability for category $c$ is defined as
% \begin{equation}
% \log p_\theta(\mathbf{r}_t \mid s_t; \mathbf{m}_t^{(c)})
% =
% \sum_{i=1}
% \mathbf{m}_t^{(c)}[i]\,
% \log \pi_\theta\!\left(
% r_{t,i} \mid \mathbf{r}_{t,<i}, s_t
% \right).
% \label{eq:masked_logprob}
% \end{equation}

% we define the PPO importance ratio for category $c$:
% \begin{equation}
% \rho_c(t)
% =
% \exp\Big(
% \log p_\theta(\mathbf{r}_t \mid s_t; \mathbf{m}_t^{(c)})
% -
% \log p_{\theta_{\text{old}}}(\mathbf{r}_t \mid s_t; \mathbf{m}_t^{(c)})
% \Big).
% \label{eq:masked_ratio}
% \end{equation}

For different tokens, we compute an independent PPO loss using its own advantage
estimate $\hat{A}_t^c$:
\begin{equation}
\mathcal{L}_c
=
\mathbb{E}_{\tau}\!\left[
\sum_t
\min\!\Big(
\rho_t^c\,\hat{A}_t^c),\;
\mathrm{clip}\big(\rho_t^c, 1-\epsilon, 1+\epsilon\big)\,\hat{A}_t^c)
\Big)
\right],
\label{eq:masked_ppo}
\end{equation}
where $c \in \{\text{sum}, \text{act}\}$, 
\(
\rho_t^c
=
\prod_{i:\mathbf{m}_t^{(c)}[i]=1}
\frac{
\pi_\theta(r_{t,i}\mid \mathbf{r}_{t,<i}, s_t)
}{
\pi_{\theta_{\text{old}}}(r_{t,i}\mid \mathbf{r}_{t,<i}, s_t)
}.
\), and $\hat{A}_t^c$ is computed from the
reward for category $c$.

\begin{figure*}[t]
  \centering
  \includegraphics[width=0.8\textwidth]{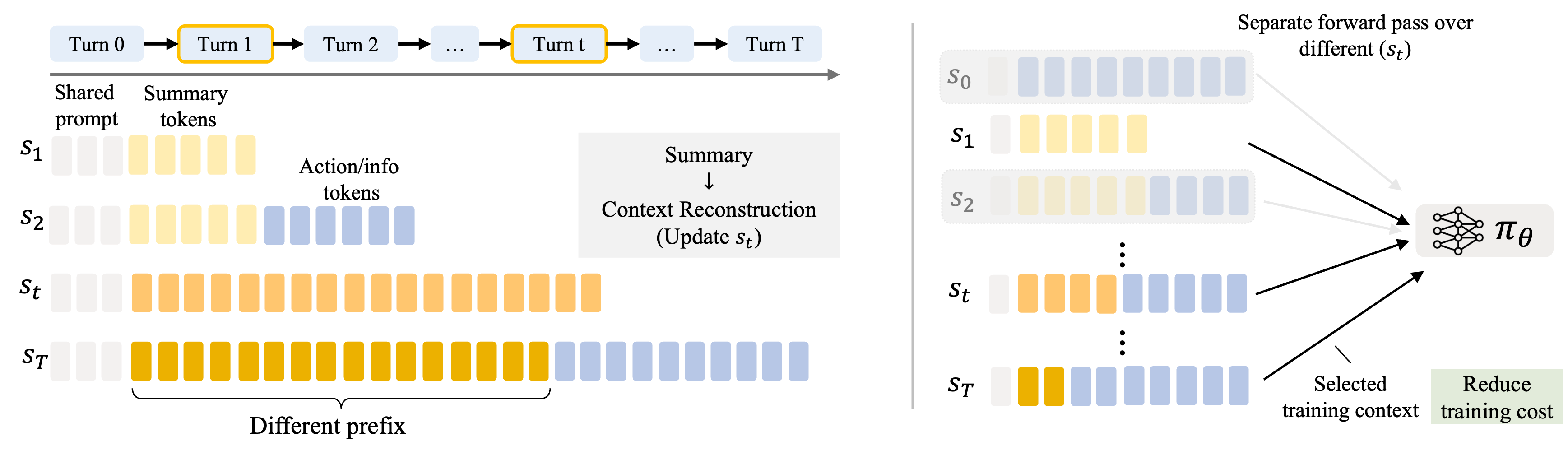}
  \caption{
    Summary-Induced Context Variation and Selective segment Training.
  Left: During rollout, each turn 
  $t$ sees a context 
  $s_t$, which evolves as the agent progresses. As summaries are generated and inserted into the context (shown in blue), they replace earlier parts of the full history (shown in yellow). This results in different turns observing different contexts.
  Right: During training, PPO requires computing 
  $\log \pi_\theta(a_t | s_t)$
  for each sampled turn. Since each $s_t$ is unique—containing distinct summary and history content—each must be fed separately into the policy network $\pi_\theta$. This prevents reuse of intermediate computations (e.g., KV cache), making training costly. To reduce this cost, we apply selective segment training, where only a subset of turns (e.g., later ones) are used for loss computation. This reduces the number of context reconstructions and forward passes, lowering training cost without changing rollout behavior.
  }
  \label{fig:selective_turn}
  \end{figure*}
  
\textbf{Summary Reward Design.} To provide proper training signals for summary generation, we design summary-specific rewards that directly evaluate the quality of folding decisions (see~\Cref{fig:summary_reward_design}). The summary reward $r_{\text{summary}}(t)$ at turn $t$ consists of two components:

\textit{Hallucination Penalty.} Summaries generated when no information is available are penalized:
\begin{equation}
r_{\text{hallucination}}(t) = \begin{cases}
-0.2 & \text{if } \text{sum}_t \text{ is generated without info} \\
0 & \text{otherwise}
\end{cases}
\label{eq:hallucination_penalty}
\end{equation}

\textit{Information Retention Reward.} Summaries that preserve critical information leading to task success are rewarded:
\begin{equation}
r_{\text{retention}}(t) = \begin{cases}
+ 0.2 & \text{if task succeeds and } \text{sum}_t \text{ was used} \\
0 & \text{otherwise}
\end{cases}
\label{eq:retention_reward}
\end{equation}

The summary reward combines both components: $r_{\text{summary}}(t) = r_{\text{hallucination}}(t) + r_{\text{retention}}(t)$. The total loss combines both components:
\begin{equation}
\mathcal{L}_{\text{total}} = \mathcal{L}_{\text{summary}} + \mathcal{L}_{\text{action}}
\label{eq:total_loss_credit}
\end{equation}
This formulation ensures that summary tokens receive independent gradient signals proportional to their importance, not their token count, enabling proper credit assignment between folding decisions and task execution actions.

\subsection{Full Context Consistency Loss}

The self-conditioning problem arises because summaries generated by $\pi_\theta$ become part of future observations, creating a feedback loop: policy updates change summary distributions $\pi_\theta(\text{sum}_t | s_t)$, which change observation distributions $p(s_t | \pi_\theta)$, violating PPO's stationary observation assumption. This leads to distribution shift between training and inference, causing unstable training signals and incorrect credit assignment.

\textbf{The Distribution Shift Problem.} In standard RL, observations follow a fixed distribution $p(s_t)$ independent of the policy. However, when summaries are generated by $\pi_\theta$ itself, the observation distribution becomes policy-dependent:
\begin{equation}
p(s_t | \pi_\theta) \neq p(s_t | \pi_{\theta_{\text{old}}}) \quad \text{(non-stationary)}
\label{eq:nonstationary_obs}
\end{equation}
This occurs because $s_t$ contains summaries $\text{sum}_t \sim \pi_\theta(\cdot | s_t)$ that are generated by the same policy model. When $\theta$ is updated, the summary distribution changes, which changes the observation distribution, creating a vicious cycle as illustrated in \Cref{fig:context_folding_overview} (\textit{Right Bottom}).

The importance sampling ratio used in PPO becomes biased:
\begin{equation}
\mathbb{E}_{s_t \sim p(\cdot | \pi_\theta)}[\rho_t(\theta)] \neq \mathbb{E}_{s_t \sim p(\cdot | \pi_{\theta_{\text{old}}})}[\rho_t(\theta)]
\label{eq:ratio_mismatch}
\end{equation}
where $\rho_t(\theta) = \frac{\pi_\theta(a_t | s_t)}{\pi_{\theta_{\text{old}}}(a_t | s_t)}$. This breaks the unbiasedness guarantee of policy gradient methods.

\textbf{Full Context Consistency Loss.} To mitigate this distribution shift, we introduce a consistency loss that encourages the policy to generate similar token distributions under both compressed context $s_t$ (used during rollout/inference) and full context $h_{0:t}$ (stored offline). During rollout, tokens $\mathbf{r}_t$ are generated using the compressed context $s_t$, while the full history $h_{0:t} = (s_0, a_0, o_1, \ldots, a_{t-1}, o_t)$ is stored for computing the consistency loss.

The consistency loss is defined as the KL divergence between policy distributions under compressed and full contexts, evaluated on the generated tokens:
\begin{equation}
\mathcal{L}_{\text{consistency}} = \mathbb{E}_\tau \left[
\sum_{t} \text{KL}\left( \pi_\theta(\cdot | s_t) \| \pi_\theta(\cdot | h_{0:t}) \right)
\right] 
\label{eq:consistency_loss}
\end{equation}
where:
\begin{itemize}[noitemsep,topsep=0pt,parsep=0pt,partopsep=0pt]
\item $\mathbf{r}_t$ are tokens generated during rollout using compressed context $s_t$ (as in training/inference)
\item $s_t$ is the compressed visible state (containing $s_0$ and the latest summary if one has been generated)
\item $h_{0:t}$ is the full interaction history (stored during rollout)
\end{itemize}

Importantly,  we only compute the log-probabilities of the \textbf{already generated tokens} $\mathbf{r}_t$ under the full context $h_{0:t}$, requiring only a single forward pass to evaluate $\log \pi_\theta(r_{t,i} | \mathbf{r}_{t,<i}, h_{0:t})$ for each token. This makes the consistency loss computationally efficient while still providing a regularization signal that encourages similar behavior under compressed and full contexts.

By minimizing the KL divergence between $\pi_\theta(\cdot | s_t)$ and $\pi_\theta(\cdot | h_{0:t})$, the consistency loss regularizes the policy and stabilizes training under the non-stationary, policy-dependent observation distribution.

Overall, the total loss is defined as:
\begin{equation}
\mathcal{L}_{\text{total}} = \mathcal{L}_{\text{summary}} + \mathcal{L}_{\text{action}} + \mathcal{L}_{\text{consistency}}
\label{eq:total_loss_selfconditioning}
\end{equation}

\subsection{Selective Segment Training}
\label{sec:selective_training}

As illustrated in \Cref{fig:selective_turn}, a key challenge in training policies that rely on summaries is the non-stationarity of context across turns. As the trajectory unfolds, summary generation modifies the visible state by replacing parts of the raw history with compressed content. This means that each turn $t$ observes a different context $s_t$, even within the same trajectory. While rollout naturally handles this, training becomes expensive: computing segment log-probabilities demands a forward pass for each unique $s_t$, which can differ in both length and content. Since Transformer models cannot reuse computation across these variable-length inputs, the training cost scales with the number of such distinct contexts.

To ensure log-probability consistency between training and inference, we must train on the actual context $s_t$ used at each turn during rollout. However, after summary generation, the context undergoes a shift: $s_t \leftarrow [s_0, \text{sum}_t]$, replacing the accumulated history with a compressed summary. This context shift must be accounted for during training to maintain consistency. Since each $s_t$ may differ in both length and content, performing a forward pass for each unique context becomes computationally expensive.

To address this challenge, we propose \textit{selective segment training}: instead of computing the loss at every single turn, we sample a subset of turns from each trajectory based on a dropout probability $p_{\text{drop}}$, and compute losses only at these selected turns. This approach drastically reduces training cost. Empirically, we find that training remains stable and effective with subsampling of turns.

\section{Experiments}

\begin{table}[t]
  \centering
  \caption{Performance on Local RAG benchmarks (F1/EM). All models use or are trained on Qwen-2.5-7B-Instruct.}
  \scriptsize
  \label{tab:main_results}
  \begin{tabular}{lcc}
    \toprule
    \textbf{Method} & \textbf{HotpotQA} & \textbf{PopQA} \\
     & \textbf{F1/EM} & \textbf{F1/EM} \\
    \midrule
    SearchR1-7B-PPO & 37.9/21.2 & 30.9/27.5 \\
    ASearcher-local-7B & 41.5/22.7 & 29.9/25.3 \\
    DeepRes.-7B & 33.4/16.9 & 30.6/25.7 \\
    ReSearch-7B & 37.0/19.2 & 32.0/27.1 \\
    Qwen-7B & 32.8/15.7 & 26.3/21.4 \\
    Qwen-7B-Fewshot & 36.6/18.8 & 30.7/24.4 \\
    \midrule
    \textbf{FoldAct-7B w.o consistency loss} & \textbf{34.9/26.7} & \textbf{33.3/29.2} \\
    \textbf{FoldAct-7B w. consistency loss} & \textbf{38.5/29.5} & \textbf{32.9/29.0} \\
    \bottomrule
    \end{tabular}
  \end{table}

  \begin{table*}[t]
    \centering
    \caption{Performance comparison on Web Search benchmarks. $^\dagger$ Baseline results are from \citet{liu2025webexplorer}.}
    \scriptsize
    \label{tab:web_search_results}
    \begin{tabular}{lccccc}
      \toprule
      \textbf{Method} & \textbf{WebWalker} & \textbf{GAIA} & \textbf{BrowseComp-en} & \textbf{BrowseComp-zh} & \textbf{XBench-DeepSearch} \\
       & \textbf{Pass@1} & \textbf{Pass@1} & \textbf{Pass@1} & \textbf{Pass@1} & \textbf{Pass@1} \\
      \midrule
      GPT-4.1-mini & 39.5 & 32.73 & 2.1 & 8.3 & 25.4 \\
      Claude-4-Sonnet$^\dagger$ & 61.7 & 68.3 & 12.2 & 29.1 & 64.6 \\
      ASearcher-Web-QwQ (32B)$^\dagger$ & 34.3 & 52.8 & 5.2 & 15.6 & 42.1 \\
      MiroThinker-8B-DPO-v0.1$^\dagger$ & 45.7 & 46.6 & 8.7 & 13.6 & -- \\
      WebSailor-7B$^\dagger$ & - & 33.0 & 6.7 & 14.2 & 34.3 \\
      \midrule
      \textbf{FoldAct-7B $p_\text{drop} =0.5$ } & \textbf{46.1} & \textbf{45.0} & \textbf{8.3} & \textbf{15.2} & \textbf{32.9} \\
      \textbf{FoldAct-7B $p_\text{drop} =0$ } & \textbf{45.7} & \textbf{46.3} & \textbf{8.4} & \textbf{14.3} & \textbf{35.4} \\
      \bottomrule
      \end{tabular}
    \end{table*}

We evaluate our method on two distinct long-horizon scenarios: (1) \textbf{Local RAG} where agents retrieve and reason over fixed knowledge bases, and (2) \textbf{Web Search} where agents navigate dynamic web environments. These scenarios differ in information structure (structured documents vs. noisy web content) and context growth patterns, allowing us to validate the generalizability of our approach.

\subsection{Experimental Setup}
\paragraph{Models \& Datasets}
For the Local RAG scenario, we evaluate Qwen-2.5-7B-Instruct (``Base'' for training, \cite{qwen2024qwen2}), its few-shot prompted version (``Few-shot''), and state-of-the-art ``RL-trained'' agents based on Qwen-2.5-7B-Instruct, including: \textsc{Search-R1} \citep{searchr1}, \textsc{ReSearch} \citep{research}, \textsc{ASearcher} \citep{asearcher} and \textsc{DeepResearcher} \citep{deepresearcher}. We evaluate the agents on two QA benchmarks: HotpotQA~\citep{hotpotqa} and PopQA~\citep{popqa}.

For the Web Search scenario, we compare against state-of-the-art agents including \textsc{ASearcher-Web-QwQ (32B)}~\footnote{\url{https://huggingface.co/inclusionAI/ASearcher-Web-QwQ}}, \textsc{GPT-4.1-mini}, \textsc{Claude-4-Sonnet}, \textsc{MiroThinker-8B-DPO-v0.1}~\footnote{\url{https://huggingface.co/miromind-ai/MiroThinker-8B-DPO-v0.1}}, and \textsc{WebSailor-7B}~\footnote{\url{https://huggingface.co/Alibaba-NLP/WebSailor-7B}}. All agents are equipped with web browsing, Python interpreter, and website visitation tools. We evaluate on four benchmarks: \textit{WebWalker}~\citep{wu2025webwalker}, \textit{GAIA}~\citep{mialon2023gaia}, \textit{BrowseComp}~\citep{wei2025browsecomp}, and \textit{XBench-DeepSearch}~\footnote{\url{https://huggingface.co/datasets/xbench/DeepSearch}}.

\subsubsection{Implementation Details}

We use Qwen-2.5-7B-Instruct as our base model and train with FoldAct, applying separated loss computation, full context consistency loss, and selective segment training ($p=0.5$) as described in Section~\ref{sec:method}. 

\textbf{Summary Format.} We use two types of summary tags to distinguish different categories of compressed information: (1) \texttt{<think\_summary>} encapsulates reasoning steps and decision-making processes, capturing the agent's thought process during task execution, and (2) \texttt{<information\_summary>} encapsulates factual information and observations retrieved from the environment. 

\textbf{Local RAG.} For SFT, we select 200 correct multi-turn trajectories from ASearcher~\footnote{\url{https://huggingface.co/datasets/inclusionAI/ASearcher-train-data}}, with summaries generated by Qwen3-30B-A3B-Instruct. We then train with FoldAct on the same dataset. 

\textbf{Web Search.} For SFT, we collect multi-turn trajectories from Qwen3-30B-A3B-Instruct on WebExplorer training dataset~\footnote{\url{https://huggingface.co/datasets/hkust-nlp/WebExplorer-QA}}, with summaries also generated by Qwen3-30B-A3B-Instruct. We then train with FoldAct on the same training dataset.

\subsection{Main Results}

\subsubsection{Performance Comparison}

\textbf{Local RAG.} Table~\ref{tab:main_results} shows performance across all benchmarks. We evaluate on two question-answering datasets that require multi-hop reasoning and information retrieval. Our method achieves competitive performance on both datasets: on HotpotQA, we achieve F1/EM scores of 38.5/29.5, with the EM score (29.5) significantly outperforming all baseline methods (the best baseline ASearcher achieves 22.7 EM). On PopQA, our method achieves F1/EM scores of 32.9/28.8, outperforming all baselines including ReSearch (32.0/27.1) and ASearcher (29.9/25.3). The improved EM scores demonstrate that our context folding approach enables more accurate answer extraction, particularly benefiting from the stabilized training and better credit assignment between summary and action tokens.

\textbf{Web Search.} Table~\ref{tab:web_search_results} compares our method against state-of-the-art agents on four web search benchmarks. FoldAct demonstrates competitive performance across all benchmarks. Compared to model size, FoldAct-7B achieves 46.1 on WebWalker, outperforming the larger ASearcher-Web-QwQ (32B) model, and also outperforms GPT-4.1-mini and MiroThinker-8B.

\begin{figure*}[t]
  \centering
  \begin{subfigure}[b]{0.48\textwidth}
    \centering
    \includegraphics[width=\textwidth]{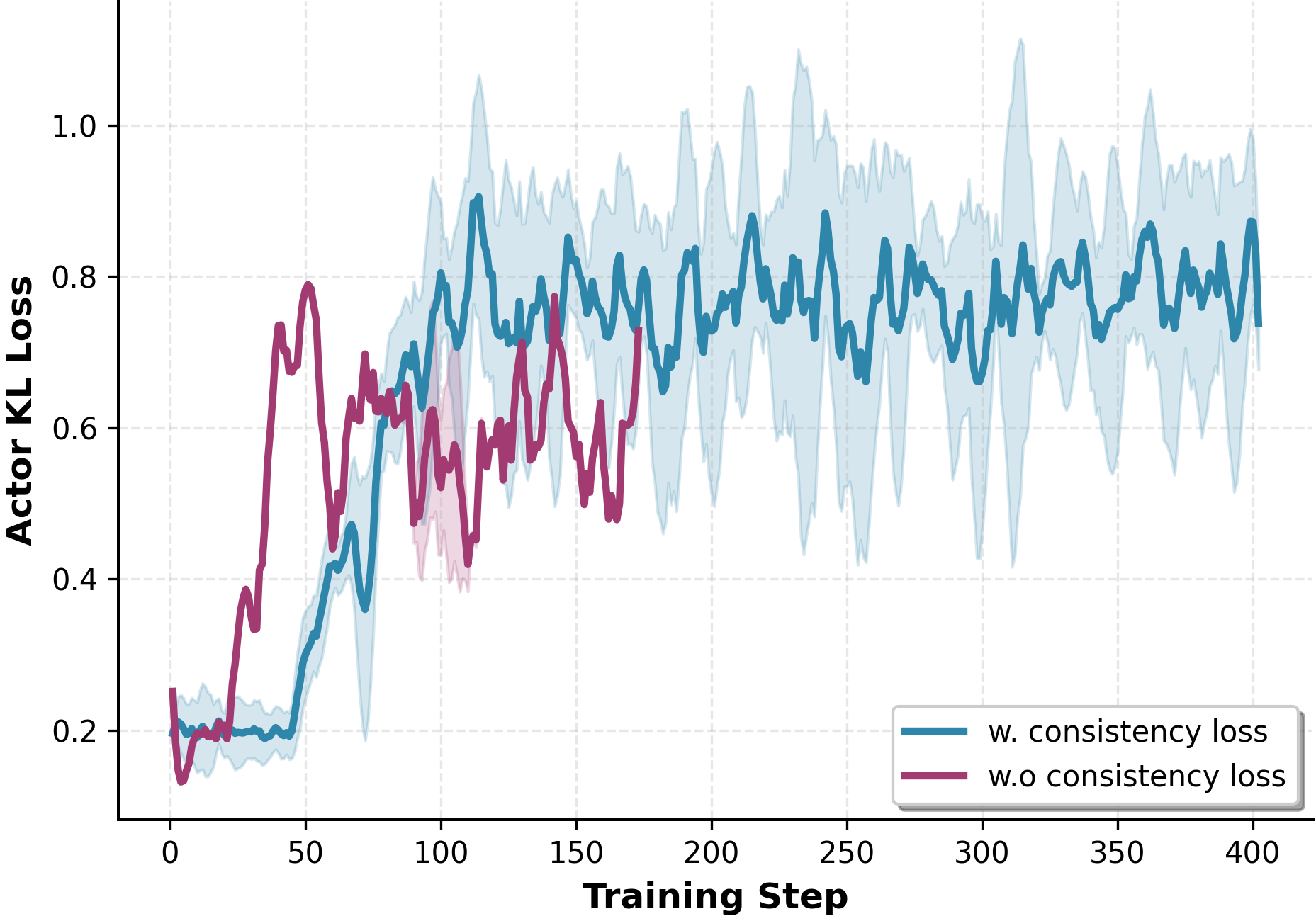}
    \caption{Actor KL Loss trends during training with and without consistency loss. FoldAct without consistency loss (purple) exhibits unstable KL loss and training collapse at step 173. In contrast, FoldAct with consistency loss (blue), the KL loss remains stable throughout training.}
    \label{fig:kl_divergence}
  \end{subfigure}
  \hfill
  \begin{subfigure}[b]{0.48\textwidth}
    \centering
    \includegraphics[width=\textwidth]{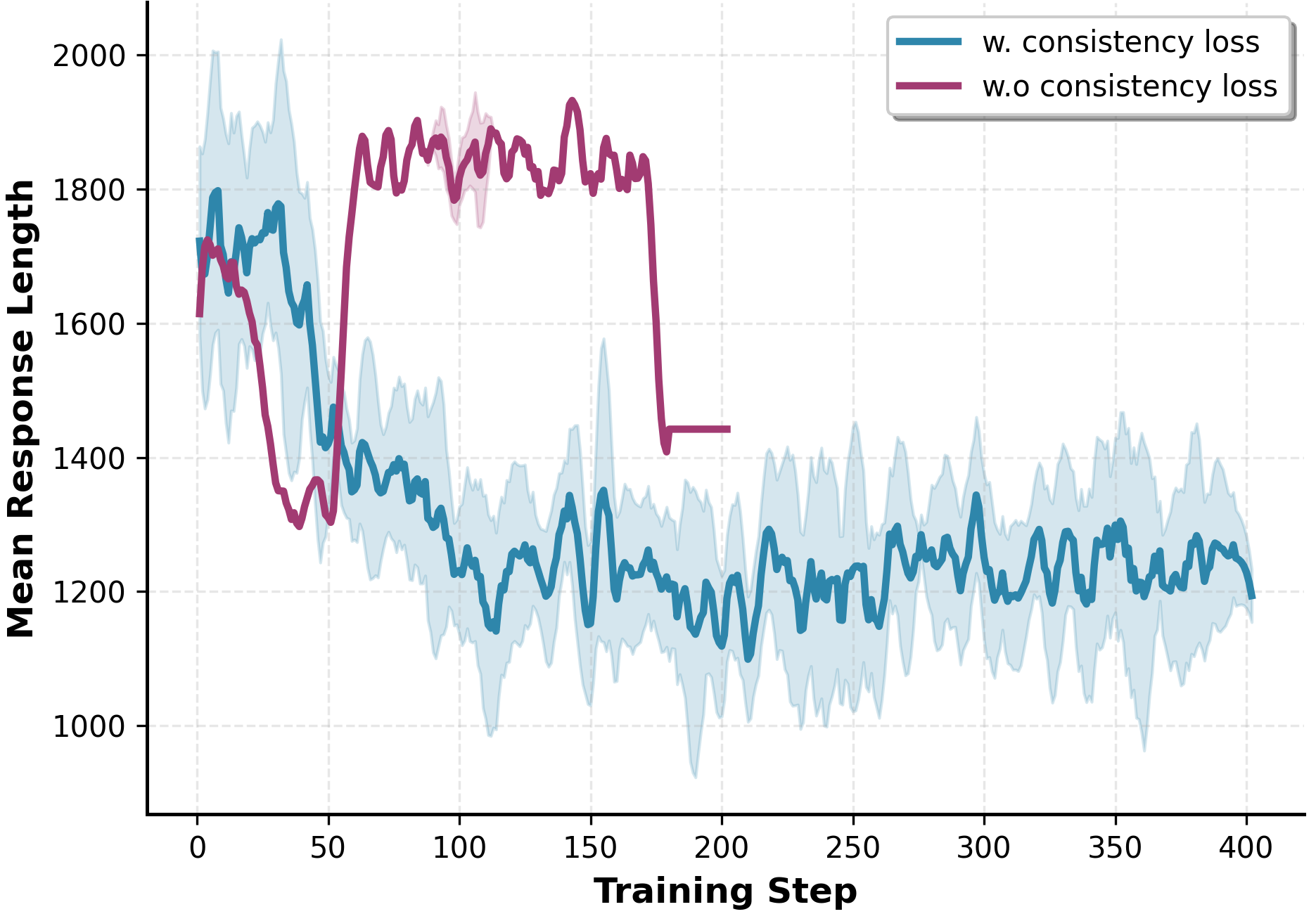}
    \caption{Response length with and without consistency loss. FoldAct with consistency loss (blue) achieves concise and stable response lengths compared to that without consistency loss (purple). After step 50, the model without consistency loss generates repetitive massive tokens, leading to training collapse.}
    \label{fig:response_length}
  \end{subfigure}
  \caption{Training stability analysis: (a) KL loss and (b) response length with and without consistency loss.}
\end{figure*}

\subsubsection{Training Stability Analysis}

Figure~\ref{fig:kl_divergence} compares the actor KL loss during training \textbf{with} and \textbf{without} the consistency loss. When consistency loss is enabled, the KL loss remains stable throughout training. In contrast, without the consistency loss, the KL loss is unstable due to distributional shift between the compressed and full contexts. Moreover, the \textit{training collapse happens} at step 173. The results highlight the critical role of consistency loss in stabilizing training and mitigating distribution shift in long-horizon scenarios with context folding.

Moreover, 
Figure~\ref{fig:response_length} shows the impact of consistency loss on response length during training. FoldAct with consistency loss achieves shorter mean response lengths (approximately 1200--1400 tokens) compared to without consistency loss, demonstrating that the consistency loss encourages more \textbf{concise and efficient responses}. However, after step 50, the model without consistency loss generates excessively repetitive massive tokens, leading to training collapse. Additionally, FoldAct with consistency loss exhibits more stable behavior.
Specifically, Table~\ref{tab:summary_example} shows a concrete example: with consistency loss, summaries are \textcolor{blue!70}{concise and focused}, while without consistency loss, summaries include \textcolor{red!70}{unnecessary process details and redundancy}.

\begin{table*}[t]
  \centering
  \caption{Summary comparison example: FoldAct with consistency loss (left) produces \textcolor{blue!70}{concise, focused summaries} (1 \texttt{think\_summary} point, essential info only), while without consistency loss (right) includes \textcolor{red!70}{redundant process details} (2--3 points, search history).}
  \scriptsize
  \label{tab:summary_example}
  \begin{tabular}{p{0.48\linewidth}|p{0.48\linewidth}}
  \toprule
  \cellcolor{blue!15}\textbf{w. consistency loss} & \cellcolor{red!15}\textbf{w.o consistency loss} \\
  \midrule
  \cellcolor{blue!5}\texttt{<think\_summary>} \newline
  \cellcolor{blue!5}1. I need to find out when Guns N Roses performed a promo for a movie starring Arnold Schwarzenegger as a former New York Police detective. \newline
  \cellcolor{blue!5}\texttt{</think\_summary>} \newline
  \cellcolor{blue!5}~ \newline
  \cellcolor{blue!5}\texttt{<information\_summary>} \newline
  \cellcolor{blue!5}The movie is called ``End of Days'' and it starred Arnold Schwarzenegger as a former New York Police Department detective named Jericho Cane. Guns N' Roses performed a promotional song for this movie during the editing phase. \newline
  \cellcolor{blue!5}\texttt{</information\_summary>} &
  \cellcolor{red!5}\texttt{<think\_summary>} \newline
  \cellcolor{red!5}1. I need to find out the year when Guns N' Roses performed a promo for a movie starring Arnold Schwarzenegger as a former New York Police detective. \newline
  \cellcolor{red!5}\textcolor{red!70}{2. I searched for information about ... and found details about his filmography and a movie called ``End of Days.''} \newline
  \cellcolor{red!5}\texttt{</think\_summary>} \newline
  \cellcolor{red!5}~ \newline
  \cellcolor{red!5}\texttt{<information\_summary>} \newline
  \cellcolor{red!5}1. Arnold Schwarzenegger starred in ``End of Days'', ... \newline
  \cellcolor{red!5}2. The film follows former New York Police Department detective Jericho Cane (played by Schwarzenegger) ... \newline
  \cellcolor{red!5}\textcolor{red!70}{3. ``End of Days'' had soundtrack songs, including ``Oh My God'' ...} \newline
  \cellcolor{red!5}\texttt{</information\_summary>}\\
  \bottomrule
  \end{tabular}
  \end{table*}
  
\subsubsection{Cost Analysis}

Due to context folding, the policy must be trained on individual trajectory segments, which increases computational cost by requiring separate forward passes for every unique $s_t$ during training.

\begin{table*}[t]
  \centering
  \caption{Training cost comparison on time per step, peak memory usage, memory reduction, and speedup. Experiments are conducted with 16$\times$NVIDIA L20 GPUs.}
  \scriptsize
  \label{tab:training_cost}
  \begin{tabular}{lcccc}
  \toprule
  \textbf{Training Strategy} & \textbf{Time/Step (s)} & \textbf{Peak Memory (GB)} & \textbf{Memory Reduction} & \textbf{Speedup} \\
  \midrule
  Full Context Training & $\sim$4846.72 & $>$441.04 (OOM) & -- & -- \\
  Ours ($p=0.5$, w. consistency) & $\sim$933.70 & 405.85 & 8.0\% & 5.19$\times$ \\
  Ours ($p=0.5$, no consistency) & $\sim$97.75 & 84.90 & 80.7\% & 49.6$\times$ \\
  \bottomrule
  \end{tabular}
  \end{table*}
  
\textbf{Cost Analysis.} Table~\ref{tab:training_cost} compares training efficiency across three strategies. \textbf{Full Context Training} processes the complete interaction history at every turn, leading to out-of-memory (OOM) errors with peak memory usage of 441.04 GB, rendering this approach infeasible for long-horizon tasks. Our \textbf{selective segment training} with $p=0.5$ (training on 50\% of turns) dramatically reduces both time and memory: without consistency loss, we achieve a 49.6$\times$ speedup (97.75s vs. 4846.72s per step) and 80.7\% memory reduction (84.90 GB vs. 441.04 GB), making training feasible. The version \textbf{with consistency loss} incurs additional cost (933.70s per step, 5.19$\times$ speedup) due to the extra forward pass required to compute $\log \pi_\theta(\cdot | h_{0:t})$ for the full context, but this cost remains manageable and is necessary for training stability.

\begin{table}[t]
\centering
\caption{Context compression efficiency analysis. We report average context length per turn (Avg Ctx.~Len./Turn) and compression ratio for Local RAG and Web Search scenarios across different number of turns.}
\scriptsize
\label{tab:compression_efficiency}
\begin{tabular}{l@{\hskip 0.7em}c@{\hskip 0.7em}r@{\hskip 0.8em}c}
\toprule
\textbf{Scenario} & \textbf{Traj.~Length} & \textbf{Avg Ctx.~Len./Turn} & \textbf{Compression Ratio} \\
    &           & \textbf{(tokens)}     & \\
\midrule
Local RAG  & 1--10 turns        & 1,892 & 0.65 \\
\midrule
\multirow{3}{*}{Web Search} 
           & 1--5 turns         & 3,986 & 0.42 \\
           & 5--10 turns        & 5,311 & 0.31 \\
           & 10+ turns          & 3,211 & 0.25 \\
\bottomrule
\end{tabular}
\end{table}

\textbf{Compression Efficiency Analysis.} Table~\ref{tab:compression_efficiency} analyzes the effectiveness of context folding across different trajectory lengths for both Local RAG and Web Search scenarios. For \textbf{Local RAG}, the compression mechanism achieves a compression ratio of 0.65 for trajectories of 1--10 turns, with an average context length per turn of 1,892 tokens. This demonstrates effective compression for structured knowledge base queries, where the relatively stable information structure allows for consistent compression performance. For \textbf{Web Search}, which involves more dynamic and noisy content, the compression behavior shows: short trajectories (1--5 turns) achieve a compression ratio of 0.42 with 3,986 tokens per turn, while medium-length trajectories (5--10 turns) show increased context length (5,311 tokens) but improved compression ratio (0.31), indicating that the folding mechanism becomes more effective as more history accumulates. For long trajectories (10+ turns), the compression ratio reaches 0.25 with a reduced context length of 3,211 tokens per turn, demonstrating that our method successfully prevents unbounded context growth even in complex web navigation scenarios. The lower compression ratios for Web Search (0.25--0.42) compared to Local RAG (0.65) reflect the need for more compression when dealing with longer, more complex web navigation trajectories.

% \begin{table*}[t]
% \centering
% \caption{Ablation study results. We report F1/EM scores on HotpotQA and NQ-Search when removing individual components. ``Full'' denotes our complete method.}
% \scriptsize
% \label{tab:ablation}
% \resizebox{\textwidth}{!}{%
% \begin{tabular}{lcc}
% \toprule
% \textbf{Method} & \textbf{HotpotQA F1/EM} & \textbf{NQ-Search F1/EM} \\
% \midrule
% Full Method & --/-- & --/-- \\
% \hline
% w/o Separated Loss & --/-- & --/-- \\
% w/o Consistency Loss & --/-- & --/-- \\
% w/o Selective Training & --/-- & --/-- \\
% \hline
% w/o Separated + Consistency & --/-- & --/-- \\
% w/o Separated + Selective & --/-- & --/-- \\
% w/o Consistency + Selective & --/-- & --/-- \\
% \bottomrule
% \end{tabular}%
% }
% \end{table*}

\section{Conclusion}

We identified a fundamental limitation in existing context folding methods for long-horizon reinforcement learning: treating summary actions as standard actions overlooks that summaries create policy-dependent, non-stationary observation distributions, violating core RL assumptions. 
We introduced \textbf{FoldAct} to enable stable and efficient training of long-horizon RL agents with context folding by separated loss computation, full context consistency loss, and selective segment training. 
Our method achieves competitive performance over state-of-the-art baselines in Local RAG and Web Search scenarios. 
With selective segment training, we achieve up to 5.19$\times$ training speedup. Moreover, the consistency loss effectively reduces distribution shift and enables more concise, efficient responses.

\newpage
\bibliographystyle{unsrtnat}
\bibliography{ref}

\end{document}